\documentclass[]{template}
\usepackage{multirow}
\usepackage{array}
\usepackage{float}
\usepackage{placeins}
\usepackage{wrapfig}
\usepackage{tikz}
\usepackage{graphicx}
\usepackage{booktabs}
\usepackage{colortbl}
\usepackage{tabularx}
\usetikzlibrary{arrows.meta,positioning,calc}
\usepackage{fontawesome5}
\usepackage{listings}
\usepackage{ragged2e}
\newcolumntype{L}{>{\RaggedRight\arraybackslash}X}
\definecolor{decisionblue}{HTML}{285B8F}
\definecolor{decisionlight}{HTML}{EEF3F8}
\hypersetup{colorlinks=true,linkcolor=compassblue,citecolor=compassblue,urlcolor=compassblue,
 pdftitle={Intern-Decision: Efficient Multimodal Decisions with Autoregressive Masked Language Modeling},
 pdfauthor={Tong Zhu et al.}}
\makeatletter
\patchcmd{\maketitle}{\rule{\textwidth}{1pt}}{\rule{\linewidth}{1pt}}{}{}
\makeatother
\renewcommand{\arraystretch}{1.16}
\setboolean{logo}{true}
\title{Intern-Decision: Efficient Multimodal Decision Making with Autoregressive Masked Language Modeling}
\newcommand{\corr}{\textcolor{black}{\fontsize{6.4}{6.4}\selectfont\faEnvelope[regular]}}
\author[1\,\protect\corr]{Tong~Zhu}
\author[1]{Zhe~Xu}
\author[1,2]{Zefeng~He}
\author[1,2]{Yunkai~Chai}
\author[1]{Xuyang~Hu}
\author[1,3]{Dehong~Zheng}
\author[1]{Yun~Luo}
\author[1]{Yafu~Li}
\author[1,3]{Qi~Zhang}
\author[1,4\,\protect\corr]{Yu~Cheng}

\affil[1]{Shanghai AI Lab}
\affil[2]{Shanghai Jiao Tong University}
\affil[3]{Fudan University}
\affil[4]{Nanyang Technological University}

\newcommand{\mailmark}{\textcolor{black}{\raisebox{0.05em}{\faEnvelope[regular]}}\hspace{0.28em}}
\newcommand{\authoremails}{%
  \normalfont\fontsize{8}{11.4}\selectfont\color{black}%
  \mbox{\mailmark\texttt{zhutong@pjlab.org.cn}}%
  \hspace{1.1em}%
  \mbox{\mailmark\texttt{chengyu@pjlab.org.cn}}%
}
\makeatletter
\patchcmd{\maketitle}{%
{\raggedright \@author\par}
                    \vskip2pt
}{%
{\raggedright \@author\par}
\par\vspace{3pt}%
{\raggedright\authoremails\par}
\vskip3pt
}{}{\PackageWarning{main}{Could not insert author emails into maketitle}}
\makeatother

\correspondingauthor{}

\begin{abstract}
Interactive agents repeatedly choose among a finite set of actions, labels, or ordinal outcomes. Generating a free-form response for each choice adds latency and complicates the interface between a model and its environment. We present Intern-Decision, a family of 0.8B, 2B, and 4B multimodal decision models that return structured choices and candidate probabilities from text and optional images. Our autoregressive masked language modeling objective places one decision placeholder per field in a complete assistant JSON skeleton. Ground-truth symbols appear only in supervision, while all fields are evaluated in a single causal forward pass using the backbone's native language-model head. The 4B model achieves a 90.02\% unweighted average accuracy across seven evaluation suites, compared with 88.74\% for Jev v1.13.0. For a fixed 289-token request with three fields on one RTX 4090, mean latency is 44.16\,ms for the 4B model and approximately 33--34\,ms for the smaller models. On a separate 96-case known-distribution pilot, temperature scaling reduces expected multiclass Brier from 0.628 to 0.550 and expected calibration error from 0.213 to 0.089. A confidence-based handoff study on 111 public Jevbench-Hard items improves accuracy from 82/111 (73.87\%) to 97/111 (87.39\%) by routing 42 items to Atria-Dawn-Preview. We also describe qualitative game, visual interface, and browser demonstrations. These results support compact decision models as a practical component of larger agent systems.
\end{abstract}

\begin{document}
\maketitle

\par\addvspace{1.2\baselineskip}
\noindent\begin{minipage}{\linewidth}
\small\RaggedRight
\faGithub\hspace{0.45em}
Code: \url{https://github.com/InternLM/Intern-Decision}\\[0.5ex]
\raisebox{-0.2ex}{\includegraphics[height=1em]{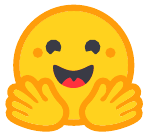}}\hspace{0.45em}
Models: \url{https://huggingface.co/collections/internlm/intern-decision}
\end{minipage}
\par\addvspace{0.8\baselineskip}

\section{Introduction}

Driven by the rapid progress of large language models (LLMs), agent systems have evolved from passive text generation models into interactive systems capable of perceiving environments, using tools, and taking actions~\cite{yao2023react,toolace2025,zhou2024webarena}.
Unlike open-ended text generation, many agent operations are inherently closed-set decision problems, where the model selects from a predefined set of alternatives, such as tool invocation, model routing, and actions~\cite{liu2023agentbench}.
However, current agents often rely on autoregressive free-form generation even when the underlying task only requires structured decisions, introducing additional decoding overhead and complicating the interface between models and downstream controllers~\cite{yao2023react,toolace2025,schick2023toolformer}.
These observations motivate the development of dedicated decision models that can directly produce structured outcomes together with reliable confidence estimates for efficient agent execution.

Recent work on structured decision models, referred to as System-1 decision models, explores this perspective by replacing unrestricted generation with explicit decision interfaces and probabilistic outputs.
Systems such as Jev~\cite{jev} provide typed interfaces for choice, score, and binary decisions, while several open implementations investigate alternative architectures and inference strategies for efficient structured decision making~\cite{jevk5,kev,semif,laya}.
These interfaces primarily target textual decisions, and implementations differ in whether they score multiple questions jointly or separately.
For multimodal agents operating in dynamic environments, decision models must additionally integrate visual observations, task instructions, and changing interface states, where action selection depends jointly on visual evidence and contextual constraints~\cite{uitars22025,uivenus152026}.
Moreover, practical agent requests often require multiple decisions over a shared state, creating a need for efficient multi-field decision models that can preserve the capabilities of pretrained multimodal language models while providing structured outputs.

To address these challenges, we introduce \textbf{Intern-Decision}, a family of  multimodal decision models based on \emph{autoregressive masked language modeling}.
Our approach reformulates structured decision making as next-token prediction over masked decision fields within a predefined assistant output schema.
By retaining the pretrained causal language model backbone and native language-model head, Intern-Decision evaluates multiple candidate decisions in a single forward pass and produces both structured predictions and candidate probabilities without iterative text generation.
This design separates decision scoring from free-form response decoding, enabling efficient inference while preserving compatibility with existing multimodal language models.

To adapt pretrained models to this decision-oriented interface, we construct a unified multimodal decision corpus consisting of textual states, optional images, typed questions, and target decisions.
The corpus combines teacher-constructed challenging decisions with public task distributions, covering textual reasoning, instruction-following decisions, visual judgments, interface actions, and interactive control scenarios.
We instantiate Intern-Decision with 0.8B, 2B, and 4B Qwen3.5 backbones~\cite{qwen35}, fine-tuning the language backbone while keeping the vision tower and projector frozen.
To improve the reliability of confidence estimates, we further apply post-hoc temperature scaling~\cite{guo2017calibration} using a separate calibration split.

We evaluate Intern-Decision from three perspectives: decision accuracy, inference efficiency, and probability quality.
Across seven evaluation suites, including public Jevbench subsets~\cite{jevbench}, Intern-Decision-4B achieves a 90.02\% unweighted mean accuracy, compared with 88.74\% for Jev.
For a fixed 289-token request containing three decision fields, Intern-Decision-4B achieves a mean latency of 44.16\,ms on a single RTX 4090.
We further evaluate probability calibration through a separate known-distribution pilot and study confidence-based model routing, in which requests are passed to a larger model when the decision model's confidence falls below a threshold.
Qualitative demonstrations on visual interaction and browser assistance further illustrate how compact decision models can serve as local decision modules within hierarchical agent systems.

Our contributions are summarized as follows:
\begin{itemize}
    \item We introduce \textbf{Intern-Decision}, a family of compact multimodal decision models that formulate structured decision making as next-token-aligned masked prediction, enabling multi-field candidate scoring with a native language-model head in a single causal forward pass.
    
    \item We develop a unified multimodal decision framework, evaluate accuracy, latency, and calibration under separate protocols, and illustrate visual interaction through qualitative demonstrations.
    
    \item We investigate confidence-based model handoff that passes low-confidence requests to a larger model, and demonstrate how compact structured decision models can complement larger models in interactive agent systems.
\end{itemize}

\section{Related Work}
\paragraph{Decision Models.}
Recent decision models target fast, structured judgments: given a state, questions, and predefined answer options, they return typed decisions with probability estimates rather than free-form text.
Alongside TypeSafe's Jev~\cite{jev}, which supports choice, score, and binary decisions and reports using Reinforcement Learning for Calibrated Decisions (RLCD), related open implementations explore different modeling and inference strategies for structured decision-making.
JevK5~\cite{jevk5} adapts Qwen3.5 with LoRA and derives candidate probabilities from next-token logits through a Jev-style interface.
Kev~\cite{kev} combines Qwen backbones with LoRA adapters and a learned pointer head, providing self-hostable models and an API for typed choices and scores.
SemIf-OpenJev~\cite{semif} provides a direct-scoring approach that derives option probabilities from candidate-token logits of frozen language models, with optional shared-state prefill reuse across questions.
Laya~\cite{laya} adopts an encoder-based, non-autoregressive architecture to produce choice, score, and yes/no decisions in a single forward pass.
Within this paradigm, Intern-Decision focuses on multimodal structured decision-making, incorporating visual inputs alongside textual states and task instructions.

\paragraph{Autoregressive and Masked Prediction.}
Autoregressive language models such as GPT-3~\cite{gpt3} predict tokens conditioned on a causal prefix, whereas BERT~\cite{bert} reconstructs masked tokens using bidirectional context.
Hybrid approaches explore combinations of these paradigms: UniLMv2~\cite{unilmv2} combines autoencoding and partially autoregressive objectives through pseudo masking, while GLM~\cite{glm} combines bidirectional encoding of the corrupted input with autoregressive reconstruction of masked spans.
Mask-Predict~\cite{maskpredict} predicts masked target tokens in parallel and iteratively refines uncertain predictions.
For classification, PET~\cite{pet} uses pattern--verbalizer pairs for cloze-based prediction and distills the resulting models into a sequence classifier, while LM-BFF~\cite{lmbff} performs prompt-based fine-tuning with label words and demonstrations, reusing the pretrained masked-language-model head.
Intern-Decision formulates structured decisions as \emph{autoregressive masked language modeling}, combining answer placeholders with next-token-aligned prediction in a causal multimodal model.
Its native language-model head scores candidates for all requested fields in a single forward pass, without iterative refinement or answer feedback.

\paragraph{Calibration and selective computation.}
Temperature scaling fits a positive scalar to adjust predictive confidence while preserving the most likely label~\citep{guo2017calibration}. Brier scoring evaluates the full predictive distribution~\citep{brier1950}, complementing accuracy and binned calibration error. Confidence-based routing and cascading use uncertainty to allocate requests between models with different costs~\citep{confidencetokens2025,cascaderouting2025}. Our evaluation separates hard-label calibration from an exact-distribution pilot and reports one escalation operating point rather than an optimized routing policy.

\section{Autoregressive Masked Decision Modeling}
\label{sec:method}
\subsection{Problem formulation}
Let $x=(s,I,Q)$ denote a request with state $s$, optional ordered images $I$, and ordered questions $Q=(q_1,\ldots,q_m)$. Question $q_j$ has a field identifier, an instruction, and an ordered candidate set $\mathcal{C}_j$. The model returns a categorical distribution $P_j$ and a selected label $\hat y_j\in\mathcal{C}_j$ for every field. The supported types are \texttt{choice}, \texttt{noul}, and \texttt{score}. Choice labels follow the insertion order of their criteria; Boolean labels are ordered as \texttt{no}, \texttt{yes}; ordinal labels are strings such as \texttt{0}, \texttt{1}, and \texttt{2}.

A deterministic mapping $a_j:\mathcal{C}_j\rightarrow\mathcal{V}$ assigns each option a legal single-token answer symbol from the tokenizer vocabulary $\mathcal{V}$. The release supports 62 candidate symbols; the HTTP service permits up to 16 fields per request. These are interface limits rather than claims about generalization to arbitrarily large schemas. Field and option order remain unchanged during compilation and scoring.

\subsection{Input construction and causal readout}
\begin{figure}[tbp]
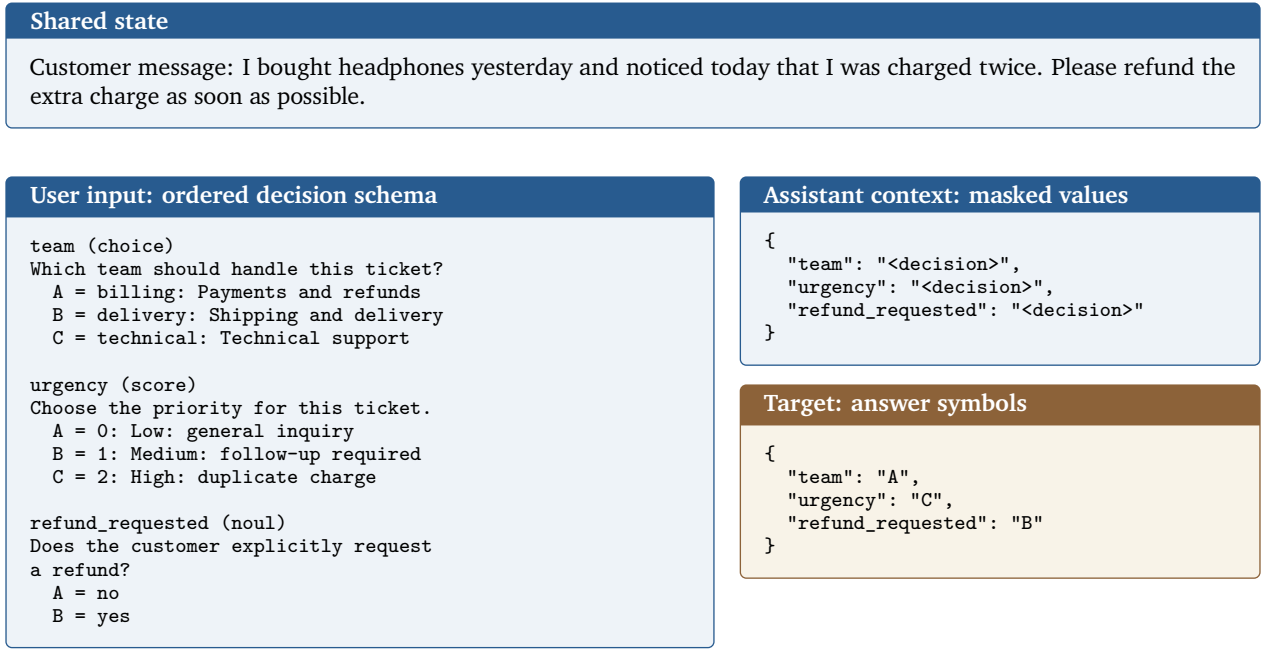

\centering
\tcbset{decisionpanel/.style={boxrule=0.5pt,arc=2pt,left=6pt,right=6pt,
  top=4pt,bottom=4pt,fonttitle=\bfseries\small,
  colback=decisionlight,colframe=decisionblue,colbacktitle=decisionblue,
  coltitle=white,nobeforeafter},
  decisioncode/.style={decisionpanel,listing only,
  listing options={basicstyle=\ttfamily\fontsize{7.2}{8.7}\selectfont,
    columns=fullflexible,keepspaces=true,breaklines=true,
    breakatwhitespace=true,breakindent=0pt,breakautoindent=false,
    showstringspaces=false,aboveskip=0pt,belowskip=0pt}}}
\begin{tcolorbox}[decisionpanel,title={Shared state}]
\small Customer message: I bought headphones yesterday and noticed today that I was charged twice. Please refund the extra charge as soon as possible.
\end{tcolorbox}
\par\vspace{6pt}
\noindent\begin{minipage}[t]{0.565\linewidth}
\vspace{0pt}
\begin{tcblisting}{decisioncode,title={User input: ordered decision schema}}
team (choice)
Which team should handle this ticket?
  A = billing: Payments and refunds
  B = delivery: Shipping and delivery
  C = technical: Technical support

urgency (score)
Choose the priority for this ticket.
  A = 0: Low: general inquiry
  B = 1: Medium: follow-up required
  C = 2: High: duplicate charge

refund_requested (noul)
Does the customer explicitly request
a refund?
  A = no
  B = yes
\end{tcblisting}
\end{minipage}\hfill
\begin{minipage}[t]{0.415\linewidth}
\vspace{0pt}
\begin{tcblisting}{decisioncode,title={Assistant context: masked values}}
{
  "team": "<decision>",
  "urgency": "<decision>",
  "refund_requested": "<decision>"
}
\end{tcblisting}
\par\vspace{6pt}
\begin{tcblisting}{decisioncode,title={Target: answer symbols},
  colback=decisionoutputbg,colframe=decisionoutputframe,
  colbacktitle=decisionoutputframe}
{
  "team": "A",
  "urgency": "C",
  "refund_requested": "B"
}
\end{tcblisting}
\end{minipage}
\caption{A constructed three-field request. Blue panels show visible evidence and masked assistant context; gold shows target answer symbols: A for billing, C for urgency 2, and B for yes. These symbols supervise the decision positions and never enter the input. System instructions, control tokens, and some option descriptions are omitted or shortened.}
\label{fig:prompt-example}
\end{figure}

The released compiler places the state and ordered decision schema in the user turn, then appends an assistant JSON skeleton with one reserved \texttt{<decision>} token per field. Figure~\ref{fig:prompt-example} presents a readable view of a constructed support-ticket request with \texttt{choice}, \texttt{score}, and \texttt{noul} fields. Choice symbols follow criteria order, score symbols follow list order from \texttt{0}, and Boolean symbols are \texttt{A} for \texttt{no} and \texttt{B} for \texttt{yes}. The target panel shows the mapped answer symbols \texttt{A}, \texttt{C}, and \texttt{B}, corresponding to \texttt{billing}, urgency \texttt{2}, and \texttt{yes}. Training supervises these symbol tokens at the decision positions; inference maps predicted symbols back to the original labels. The displayed JSON groups the supervision targets for illustration and is not inserted into the input or supervised as a complete generated JSON response.

\definecolor{decisionoutputbg}{HTML}{F8F3E8}
\definecolor{decisionoutputframe}{HTML}{8C6239}

Labels, reference distributions, benchmark rationales, and provenance are excluded from the visible evidence. Training and inference use the same masked skeleton, and changing a target must not change the input token sequence.

Let $p_j$ be the token position of the placeholder for field $j$. A causal backbone computes hidden states $h_1,\ldots,h_L$, and the native LM head produces vocabulary logits $z_t$. The released objective reads the position immediately before the marker:
\begin{equation}
 z_{p_j-1}=\operatorname{LMHead}(h_{p_j-1}).
 \label{eq:readout}
\end{equation}
The answer label is stored at position $p_j$ in the supervision tensor, so the ordinary causal shift pairs it with $z_{p_j-1}$. No ground-truth answer symbol is inserted into the input. Earlier placeholders are visible to later fields through causal attention, but their gold or predicted answers are not fed back. All requested logits are collected from the same forward pass.

\begin{figure}[htbp]
\centering
\begin{tikzpicture}[font=\small,>=Latex,
 box/.style={draw=decisionblue,rounded corners=2pt,align=center,inner sep=8pt},
 flow/.style={->,thick,draw=decisionblue}]
\node[box,text width=3.0cm] (state) at (0,0) {State and\\optional images};
\node[box,text width=3.0cm] (schema) at (4.2,0) {Ordered questions and candidate labels};
\node[box,text width=3.7cm] (mask) at (9,0) {Assistant JSON skeleton\\all values are placeholders};
\node[box,fill=decisionlight,text width=11.8cm] (model) at (4.5,-1.7) {Qwen3.5 causal backbone and native LM head\\one forward pass; no answer feedback};
\node[box,text width=4.4cm] (slots) at (1.9,-3.4) {Read logits immediately before each decision marker};
\node[box,text width=4.4cm] (outputs) at (7.5,-3.4) {Legal-option softmax at fixed $T$\\labels, probabilities, structured output};
\draw[flow] (state.south) -- (state.south|-model.north);
\draw[flow] (schema.south) -- (schema.south|-model.north);
\draw[flow] (mask.south) -- (mask.south|-model.north);
\draw[flow] (model.south) -- (slots.north);
\draw[flow] (slots.east) -- (outputs.west);
\end{tikzpicture}
\caption{Intern-Decision inference. All fields share the same state and schema, and their placeholders are processed together. The readout uses next-token alignment, not same-position supervision.}
\label{fig:method}
\end{figure}
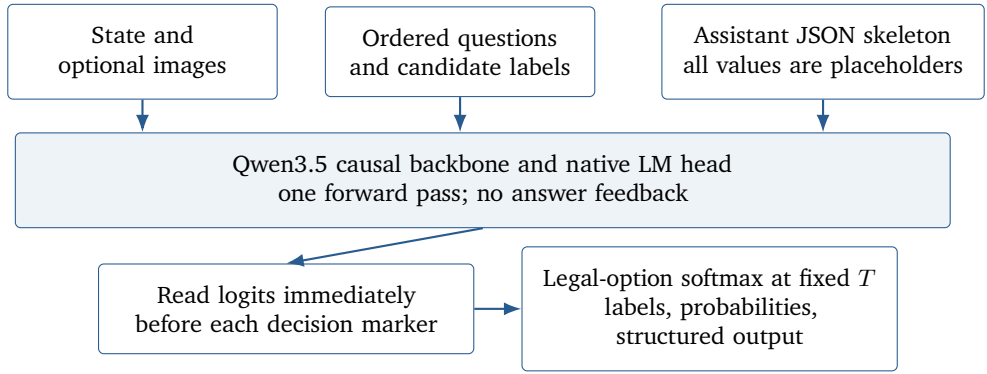

\subsection{Training objective}
For a batch with $M$ supervised fields, the loss is full-vocabulary cross-entropy at the shifted decision positions:
\begin{equation}
 \mathcal{L}_{\mathrm{decision}}
 =-\frac{1}{M}\sum_{j=1}^{M}
 \log\frac{\exp z_{p_j-1,a_j(y_j)}}
 {\sum_{v\in\mathcal{V}}\exp z_{p_j-1,v}}.
 \label{eq:loss}
\end{equation}
All other labels are ignored. Thus JSON punctuation, state tokens, and schema descriptions are context rather than supervised output targets. The objective is neither full-JSON generation loss nor full-token continued pretraining. The training path uses hard labels.

\subsection{Candidate probabilities and structured outputs}
At inference, only the legal candidate logits for each question are normalized:
\begin{equation}
 P_j(c\mid x;T)=
 \frac{\exp(z_{p_j-1,a_j(c)}/T)}
 {\sum_{c'\in\mathcal{C}_j}\exp(z_{p_j-1,a_j(c')}/T)},
 \qquad \hat y_j=\mathop{\arg\max}_{c\in\mathcal{C}_j} P_j(c\mid x;T).
 \label{eq:probability}
\end{equation}
A positive scalar $T$ changes the probability spread without changing the argmax, apart from implementation-level numerical ties. The serializer maps symbols back to original labels and creates the response object. Consequently, schema-conforming output does not depend on a free-text decoder generating valid JSON. It also does not guarantee semantic correctness or agreement between related fields.

The service can return the probability of \texttt{yes} for Boolean questions and an expected numeric value for ordinal questions. These quantities differ from the discrete argmax label used for decision accuracy. For all types, candidate probabilities and maximum probability remain available to downstream controllers.

\subsection{Computational scope}
The method removes iterative answer decoding and avoids separate field-by-field forward passes. It does not eliminate the cost of processing the input: longer states, more image tokens, and larger schemas still increase computation and memory. One request may yield multiple decisions, so request rate and decision rate are different quantities. Causal ordering also means fields can have different contextual prefixes even when they share a forward pass; the construction does not enforce permutation invariance or a jointly constrained output distribution.

\section{Data Construction}
\subsection{Decision-oriented data construction}
\label{sec:decision-data-construction}

The decision corpus teaches structured choices rather than free-form answers. It combines synthetic textual decisions, public task distributions, and synthetic visual and interactive tasks. Each example contains a visible state, optional images, typed questions, and target labels, allowing these sources to share the same decision interface.

\begin{table}[htbp]
    \centering
    \caption{Training dataset composition.}
    \label{tab:decision-data-categories}
    \small
    \renewcommand{\arraystretch}{1.18}
    \begin{tabularx}{\linewidth}{@{}>{\RaggedRight\arraybackslash}p{0.22\linewidth}LL@{}}
    \toprule
    \textbf{Component} & \textbf{Representative sources / tasks} & \textbf{Capability focus} \\
    \midrule
    Synthetic text decisions & General and Hard decisions synthesized by \texttt{deepseek-v4.1-flash} & Evidence use, rule following, and difficult textual judgments \\ \addlinespace[3pt]
    Public task distributions & AG News~\citep{agnews2015}, ToolACE~\citep{toolace2025}, Typed Decisions~\citep{typeddecisions2026}, WildGuardMix~\citep{wildguard2024}, WildJailbreak~\citep{wildteaming2024}, and Ruozhiba (LLM Ruozhiba QA and COIG-CQIA)~\citep{ruozhibaqa,coigcqia2024} & Classification, tool routing, typed decisions, safety, and reasoning \\ \addlinespace[3pt]
    Synthetic visual and interactive tasks & Game environments, VizDoom, and shapes/captcha-style tasks & Visual grounding, discrete control, and pointer-action selection \\
    \bottomrule
    \end{tabularx}
\end{table}

\begin{wrapfigure}[37]{r}{0.475\linewidth}
\centering
\includegraphics[width=\linewidth]{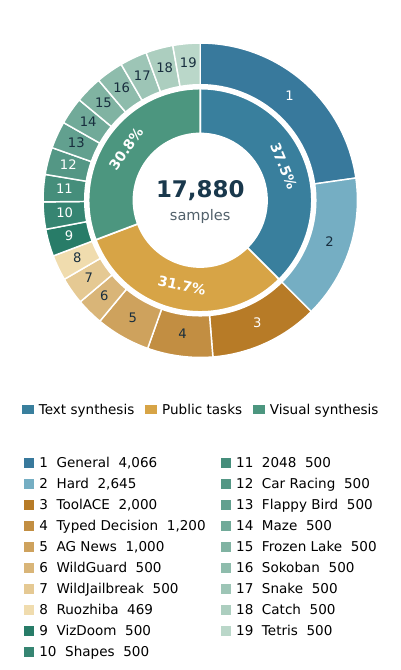}
\captionsetup{font=small,hypcap=false}
\caption{Training mixture shared by the released models. The inner ring shows components; numbered outer sectors show data categories, including each game. Counts are training-input samples.}
\label{fig:training-data-composition}
\end{wrapfigure}

The released 0.8B, 2B, and 4B checkpoints use the same training input: 17,880 samples and 23,500 decisions, including 5,500 image-bearing samples. This identity is verified against input hashes, training logs, and tokenizer caches for all three runs. Figure~\ref{fig:training-data-composition} shows three components and nineteen categories, with the nine games listed individually. Sector areas encode samples, so a multi-question record contributes once even when it supplies several supervised decisions.

\Needspace{35\baselineskip}
The synthetic text component uses \texttt{deepseek-v4.1-flash}~\footnote{\url{https://huggingface.co/deepseek-ai/DeepSeek-V4.1-Flash}} as the teacher model. As Figure~\ref{fig:training-data-composition} shows, General contains 4,066 general-category synthetic decisions; Hard combines the hard and patch batches into 2,645 difficult-category synthetic decisions. We follow the Jevbench ability type to construct such general and hard synthetic decision examples. Public sources cover news classification, tool routing, typed decisions, and safety judgments. Synthetic visual sources supply rendered game frames, VizDoom observations, and shapes-based pointer actions. Appendix~\ref{app:training-data-examples} provides component counts, source descriptions, and representative records.

The visual sources require recognition, comparison, target localization, and discrete action selection. Shapes tasks connect object attributes and spatial relations to pointer movements; game and VizDoom tasks connect frames to control decisions. These capabilities overlap within the recorded sources. Table~\ref{tab:decision-data-categories} summarizes the evidenced source groups and their capability focus.

\subsection{Calibration test set construction}
\label{sec:calibration-data}
This evaluation-only test set measures whether the model assigns appropriate probabilities to all possible outcomes of a stated random experiment. For example, a fair die requires probability $1/6$ on each real face and zero on impossible faces. Selecting one plausible face does not establish a correct predictive distribution. Similarly, the Monty Hall questions ask where the prize is, with doors as outcomes; they do not ask whether switching is the best action.

The set is constructed from 24 mechanism families in six categories. Each family has two settings, which vary numerical parameters or the observation protocol. Each setting is emitted with canonical and reversed option order, yielding $6\times4\times2\times2=96$ cases, 16 per category. The two order variants share the same state, question, and reference distribution. There are therefore 48 paired settings rather than 96 independent trials. Appendix~\ref{app:calibration-cases} gives the family inventory and concrete input/reference examples.

For each setting, a deterministic Python generator specifies the sample space, priors, independence assumptions, and observation or selection mechanism. It computes the exact distribution using rational arithmetic: counting and enumeration for finite outcomes, conditional-probability formulas for observed evidence, a forward recursion for hidden states, and combinatorial or dynamic-programming calculations for sequential processes. References are not Monte Carlo estimates or model-generated answers. Construction checks enforce probabilities in $[0,1]$, an exact sum of one, matching option keys, unique IDs, and equivalent reversed-order pairs. Rebuilding from the archived generator reproduces the evaluated input and reference files byte for byte.

Inputs contain only the state and typed questions. Exact fractions, numerical reference probabilities $q_i$, derivations, and provenance remain in a separate reference file and are never sent to the model. The set is not used to train the model, fit its temperature, or select a checkpoint. An exact-content check found no overlap with the temperature-fitting records; this check does not establish absence of semantic overlap with every training source.

In the actual three-door informed-host case, the prize is uniformly placed and the player initially selects door 1. A host who knows the prize location always opens an unchosen empty door and always offers the other unopened door; legal reveals are chosen uniformly. The input does not disclose the opened door's number. The question is ``Where is the prize relative to the final two unopened doors?'' Its outcomes are \texttt{initial\_door} and \texttt{other\_unopened\_door}, with reference probabilities $1/3$ and $2/3$. The initial selection retains its prior probability, and the host's guaranteed safe reveal concentrates the other $2/3$ on the remaining door. An output concentrated entirely on the remaining door would select the more likely outcome but misrepresent the residual uncertainty.

\section{Evaluation Results}
\label{sec:evaluation}
\subsection{Benchmark scope and scoring}
Table~\ref{tab:datasets} lists the evaluated bundles. Accuracy is computed per decision, not per row; this matters for Typed Decision, whose 400 records contain 2,000 fields. The seven-suite average is an unweighted mean of suite accuracies, so each suite contributes equally despite their different sizes. It is not pooled accuracy over all 12,351 decisions.

\begin{table}[htbp]
\centering
\caption{Accuracy evaluation scope. Counts refer to the released test bundle, not the full upstream datasets.}
\label{tab:datasets}
\begin{tabularx}{\linewidth}{LrrL}
\toprule \textbf{Suite} & \textbf{Rows} & \textbf{Decisions} & \textbf{Evaluation focus} \\ \midrule
Jevbench-Easy & 48 & 48 & Explicit facts and direct decisions \\
Jevbench-Original & 72 & 72 & Policy, routing, and typed judgments \\
Jevbench-Hard & 111 & 111 & Longer contexts and interacting rules \\
Typed Decision & 400 & 2,000 & Multiple typed fields per state \\
ToolACE & 310 & 310 & Tool selection \\
AG News & 7,600 & 7,600 & News classification \\
WildJailBreak & 2,210 & 2,210 & Safety-related classification \\ \midrule
Total & 10,751 & 12,351 & Seven separately weighted suites \\
\bottomrule
\end{tabularx}
\end{table}

The public Jevbench subset contains 231 items across Easy, Original, and Hard~\citep{jevbench}. It excludes private items and the separate judge tier. Its official leaderboard combines multiple axes; the accuracy averages in this report do not reproduce that composite. Public Jevbench has informed development and should be treated as a diagnostic, not independent held-out generalization evidence. Input identities, labels, option order, and upstream scoring definitions are preserved.

For a hard-label target $y_i$, the multiclass Brier score~\citep{brier1950} is
\begin{equation}
 \mathrm{BS}=\frac{1}{N}\sum_{i=1}^{N}\sum_{k\in\mathcal{C}_i}
 \left(p_{ik}-\mathbb{1}[k=y_i]\right)^2.
 \label{eq:brier}
\end{equation}
This is a sum over classes, including both terms for a binary question. Hard-benchmark ECE uses confidence $c_i=\max_k p_{ik}$ and 10 equal-width bins:
\begin{equation}
 \mathrm{ECE}=\sum_{b=1}^{10}\frac{|B_b|}{N}
 \left|\frac{1}{|B_b|}\sum_{i\in B_b}c_i
 -\frac{1}{|B_b|}\sum_{i\in B_b}\mathbb{1}[\hat y_i=y_i]\right|,
 \label{eq:ece}
\end{equation}
with empty bins omitted. Brier and ECE beside the accuracy results are computed only on the 111 Hard items. They are not seven-suite averages. Public Hard also has 10 exact-distribution items used for TVD by the upstream scorer; this small subset is distinct from the 96-case pilot below.

\subsection{Decision accuracy}
\begin{table}[htbp]
\centering
\caption{Decision accuracy in percent. Average is the unweighted mean of the seven suites. Denominators are given in Table~\ref{tab:datasets}; bold data values mark column maxima, including ties. The comparison is restricted to these evaluated systems and protocols.}
\label{tab:accuracy}
\begingroup\footnotesize\setlength{\tabcolsep}{3.4pt}
\begin{tabularx}{\linewidth}{@{}l*{8}{>{\raggedleft\arraybackslash}X}@{}}
\toprule
\textbf{Model} & \textbf{Easy} & \textbf{Original} & \textbf{Hard} & \textbf{Typed} & \textbf{ToolACE} & \textbf{AG News} & \textbf{WildJB} & \textbf{Average} \\ \midrule
Jev & \textbf{100.00} & \textbf{98.61} & 72.07 & 73.35 & 91.29 & 89.57 & \textbf{96.29} & 88.74 \\
Laya & 95.83 & 72.22 & 28.83 & 35.95 & 63.87 & \textbf{92.84} & 14.84 & 57.77 \\
SemIf & \textbf{100.00} & \textbf{98.61} & 61.26 & 62.80 & 85.16 & 89.22 & 92.53 & 84.23 \\
Kev & \textbf{100.00} & 93.06 & 45.05 & 65.60 & 87.42 & 89.82 & 75.97 & 79.56 \\
JevK5 & \textbf{100.00} & 97.22 & \textbf{73.87} & 64.50 & 80.97 & 89.13 & 90.45 & 85.16 \\
\rowcolor[HTML]{F0F4FF} Intern-Decision-0.8B & 97.92 & 80.56 & 52.25 & 77.35 & 94.52 & 88.61 & 64.48 & 79.38 \\
\rowcolor[HTML]{F0F4FF} Intern-Decision-2B & \textbf{100.00} & 84.72 & 63.96 & 79.35 & \textbf{96.45} & 89.96 & 78.33 & 84.68 \\
\rowcolor[HTML]{F0F4FF} Intern-Decision-4B & \textbf{100.00} & \textbf{98.61} & \textbf{73.87} & \textbf{80.55} & \textbf{96.45} & 90.82 & 89.86 & \textbf{90.02} \\
\bottomrule
\end{tabularx}
\endgroup
\end{table}

Intern-Decision-4B has a suite average of 90.02\%, exceeding the baselines in Table~\ref{tab:accuracy}, and leads or ties on five of the seven suites. It scores 1,611/2,000 (80.55\%) on Typed Decision and 299/310 (96.45\%) on ToolACE. On public Jevbench, it answers 48/48 Easy, 71/72 Original, and 82/111 Hard items correctly. Jev leads WildJailBreak, Laya leads AG News, and JevK5 ties the 4B model on Hard. These results describe the evaluated systems; without matched-data ablations, they do not isolate the effect of joint field scoring or model size.

Within Intern-Decision, the suite average rises monotonically from 79.38\% at 0.8B to 84.68\% at 2B and 90.02\% at 4B, with Typed Decision and ToolACE already comparatively strong at 0.8B. Without matched data and training-compute disclosures or ablations, this pattern should not be attributed solely to parameter count.

\begin{table}[htbp]
\centering
\caption{Jevbench-Hard probability metrics on 111 items. ECE uses maximum candidate probability and 10 equal-width bins. Intern-Decision uses separately fitted temperatures; baseline probabilities use their reported/default settings. Temperatures are not comparably refitted for every baseline.}
\label{tab:hard-calibration}
\begin{tabular}{lrrr}
\toprule \textbf{Model} & {\bfseries\boldmath Brier $\downarrow$} & {\bfseries\boldmath ECE $\downarrow$} & \textbf{Temperature} \\ \midrule
Jev & 0.358420 & 0.094685 & Reported/default \\
Laya & 0.804244 & 0.246455 & Reported/default \\
SemIf & 0.498012 & 0.112213 & 1.000000 \\
Kev & 0.738253 & 0.262200 & 2.962195 \\
JevK5 & 0.366162 & \textbf{0.046672} & 1.532000 \\
\rowcolor[HTML]{F0F4FF} Intern-Decision-0.8B & 0.529519 & 0.065710 & 2.747761 \\
\rowcolor[HTML]{F0F4FF} Intern-Decision-2B & 0.437256 & 0.100208 & 2.100509 \\
\rowcolor[HTML]{F0F4FF} Intern-Decision-4B & \textbf{0.346762} & 0.065267 & 1.992418 \\
\bottomrule
\end{tabular}
\end{table}

\begin{wrapfigure}[24]{r}{0.475\linewidth}
\centering
\includegraphics[width=\linewidth]{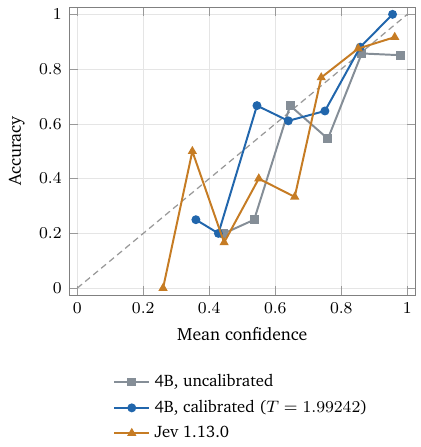}
\captionsetup{font=small,hypcap=false}
\caption{Jevbench-Hard reliability: 111 items and ten equal-width bins. Only populated bins are shown.}
\label{fig:reliability}
\end{wrapfigure}

The 4B model has the lowest Hard Brier in this comparison, whereas JevK5 has the lowest Hard ECE. Figure~\ref{fig:reliability} compares the released 4B checkpoint before and after temperature scaling with Jev~1.13.0 on the same 111 items.

Temperature scaling leaves the 4B model's 82/111 correct decisions unchanged while reducing ECE from 0.132613 to 0.065267. Jev answers 80/111 correctly and has ECE 0.094685. The temperature was fitted on separate calibration data; these Hard items were not used to choose it.

\Needspace{24\baselineskip}
Each point reports mean maximum-candidate confidence and empirical accuracy within one of ten equal-width bins. Empty bins are omitted. The curves use archived prediction statistics rather than digitized image coordinates. Connecting lines are descriptive: points summarize different bins and do not trace the same items across calibration settings.

The bins have unequal sample counts, including one Jev bin with a single item. Accordingly, the plot supports an aggregate calibration comparison but does not establish reliable behavior at every confidence level. No bin-level uncertainty intervals are available.
\subsection{Local inference latency}
Local measurements use one RTX 4090, BF16, SDPA, the native Hugging Face backend, Transformers 5.14.1, and flash-linear-attention 0.4.2. Each request has 289 input tokens and three fields: one choice, one Boolean, and one ordinal question. All three are answered in one forward pass with checkpoint-specific temperature scaling. This is a fixed short-request measurement, not a latency sweep over long contexts or screenshots.

\WFclear
\begin{table}[htbp]
\centering
\caption{Reported request latency in milliseconds. Ratios are Jev mean divided by model mean. The supplied protocol establishes the local Intern-Decision configuration; Jev's complete serving and transport conditions still require documentation.}
\label{tab:latency}
\begin{tabular}{lrrrr}
\toprule \textbf{Model} & \textbf{Mean} & \textbf{P50} & \textbf{P95} & \textbf{Mean ratio} \\ \midrule
Jev & 109.70 & 106.30 & 146.70 & 1.00$\times$ \\
\rowcolor[HTML]{F0F4FF} Intern-Decision-0.8B & 33.98 & 33.44 & 37.50 & 3.23$\times$ \\
\rowcolor[HTML]{F0F4FF} Intern-Decision-2B & 33.28 & 33.15 & 33.55 & 3.30$\times$ \\
\rowcolor[HTML]{F0F4FF} Intern-Decision-4B & 44.16 & 44.03 & 44.60 & 2.48$\times$ \\
\bottomrule
\end{tabular}
\end{table}

The reciprocal mean latencies imply approximately 29.43, 30.05, and 22.64 serial requests per second for 0.8B, 2B, and 4B, respectively, assuming negligible additional loop overhead. These are derived request rates rather than measured saturated serving throughput. They do not imply the same frame rate for an interactive agent, which also incurs observation, image preprocessing, action execution, and environment-update costs. Batch size, warm-up, repetition count, timing boundaries, and Jev transport conditions are needed before interpreting the ratios as controlled hardware speedups.

\subsection{Known-distribution calibration pilot}
\label{sec:pilot}
The known-distribution calibration test set constructed in Section~\ref{sec:calibration-data} contains 96 cases across six categories. It evaluates complete outcome distributions against exact reference probabilities.

The question wording requests an outcome, while the native decision interface returns a probability for every listed option. Intern-Decision maps the ordered options to legal answer tokens and applies the candidate softmax in Equation~\ref{eq:probability}; Jev supplies its native API probability map. We score these complete distributions, rather than converting a selected label into a one-hot vector or counting how often repeated sampled answers occur. Normalization guarantees a probability vector, but correctness must be measured against $q_i$.

\begin{table}[htbp]
\centering
\caption{Known-distribution pilot results. Each cell is expected Brier / expected ECE; bold marks the lowest value across the three settings for each metric separately. All settings have 96/96 valid predictions. Category rows have 16 cases each. Overall ECE is recomputed on the pooled cases, not averaged over categories.}
\label{tab:pilot}
\begingroup\small
\begin{tabularx}{\linewidth}{Lc*{2}{>{\columncolor[HTML]{F0F4FF}}c}}
\toprule \textbf{Category} & \textbf{Jev} & \textbf{4B uncalibrated} & \textbf{4B calibrated} \\ \midrule
Direct randomness and support & 0.490 / 0.216 & 0.483 / 0.181 & \textbf{0.421} / \textbf{0.129} \\
Composed events and mixtures & 0.682 / 0.274 & 0.677 / 0.254 & \textbf{0.577} / \textbf{0.150} \\
History, conditioning, hidden state & 0.657 / 0.113 & 0.711 / 0.219 & \textbf{0.613} / \textbf{0.108} \\
Daily evidence and observation bias & 0.603 / \textbf{0.114} & 0.701 / 0.328 & \textbf{0.575} / 0.210 \\
Selective disclosure and puzzles & \textbf{0.483} / 0.138 & 0.540 / 0.119 & 0.510 / \textbf{0.049} \\
Sequential and combinatorial processes & 0.657 / 0.116 & 0.656 / 0.180 & \textbf{0.605} / \textbf{0.058} \\ \midrule
Overall pooled & 0.595 / 0.130 & 0.628 / 0.213 & \textbf{0.550} / \textbf{0.089} \\
\bottomrule
\end{tabularx}
\endgroup
\end{table}

The final 4B checkpoint is evaluated with a fixed XTuner backend and $T=1.992418$, fitted on separate calibration records. Jev uses version 1.13.0. Temperature scaling reduces overall expected Brier by 0.078 and expected ECE by 0.124 while preserving every argmax. The calibrated model improves on its uncalibrated result in both metrics in all six categories. Relative to Jev, it has lower Brier in five categories and lower ECE in five categories; it is worse on Brier for selective disclosure and on ECE for daily evidence. Overall ECE differs from the mean category ECE because the cases are rebinned together.

For the informed-host three-door case described in Section~\ref{sec:calibration-data}, calibrated Intern-Decision returns $(0.4222,0.5778)$ and Jev returns $(0.18,0.82)$ in that order. Both select the more likely door, while their total variation distances from the reference are 0.0889 and 0.1533, respectively. The uncalibrated 4B distribution $(0.3486,0.6514)$ is closer still, so the overall gain from temperature scaling does not imply improvement on every case. Jev also returns a separate confidence of 0.63 here, which is the value used by the pilot ECE protocol. Changing the observation mechanism changes the target: an uninformed host who randomly opens a door, conditioned on revealing no prize, yields $(1/2,1/2)$. Appendix~\ref{app:calibration-cases} also explains a selective-offer variant.

\subsection{Confidence-based model handoff}
\label{sec:handoff}
Let $c(x)$ be a scalar confidence from the small model. A simple cascade retains its answer when $c(x)\geq\tau$ and invokes a larger model otherwise. The reported study routes 42 of the 111 public Hard items to Atria-Dawn-Preview~\citep{atria}, leaving 69 items with Intern-Decision-4B. The routed fraction is 37.84\%, and local coverage is 62.16\%.

\begin{table}[htbp]
\centering
\caption{Public model handoff study on Jevbench-Hard. The hybrid time is an estimate, while the Atria-only total is reported as measured. The local total is derived from 111 times the fixed-request mean latency and is not a per-item Hard timing trace.}
\label{tab:handoff}
\begin{tabularx}{\linewidth}{LrrrL}
\toprule \textbf{System} & \textbf{Correct / 111} & \textbf{Accuracy} & \textbf{Atria calls} & \textbf{Total time} \\ \midrule
\rowcolor[HTML]{F0F4FF} Intern-Decision-4B & 82 & 73.87\% & 0 & 4.90\,s, derived \\
Confidence-based handoff & 97 & 87.39\% & 42 & 140.16\,s, estimated \\
Atria-Dawn-Preview only & 99 & 89.19\% & 111 & 300.03\,s, measured \\
\bottomrule
\end{tabularx}
\end{table}

The hybrid answers 15 more items correctly than the small model and two fewer than Atria alone. Relative to the reported Atria-only total, its estimated time is 53.29\% lower. This illustrates a potentially useful operating point rather than an established optimal policy. 

  \section{Multimodal Interaction and Model Collaboration}
\subsection{A closed-loop decision interface}
In an interactive controller, each iteration observes an environment state, forms candidate actions, evaluates them, executes the selected action, and observes the result. Intern-Decision supplies the local choice within this loop. The caller defines the action space and determines when to request another decision. The following demonstrations illustrate this interface; they are selected qualitative trajectories rather than controlled benchmark measurements.

\subsection{Visual game control}
Five described demonstrations span Snake, fruit catching, Mario, and two first-person Doom tasks. They require different local decisions: selecting a safe direction toward food, distinguishing fruit from hazards, coordinating movement and jumping, searching for health, or targeting an enemy. Screenshots provide visual evidence directly, so the controller can operate without converting every game state into a symbolic text description.

\begin{figure}[!htb]
\centering
\includegraphics[width=\linewidth]{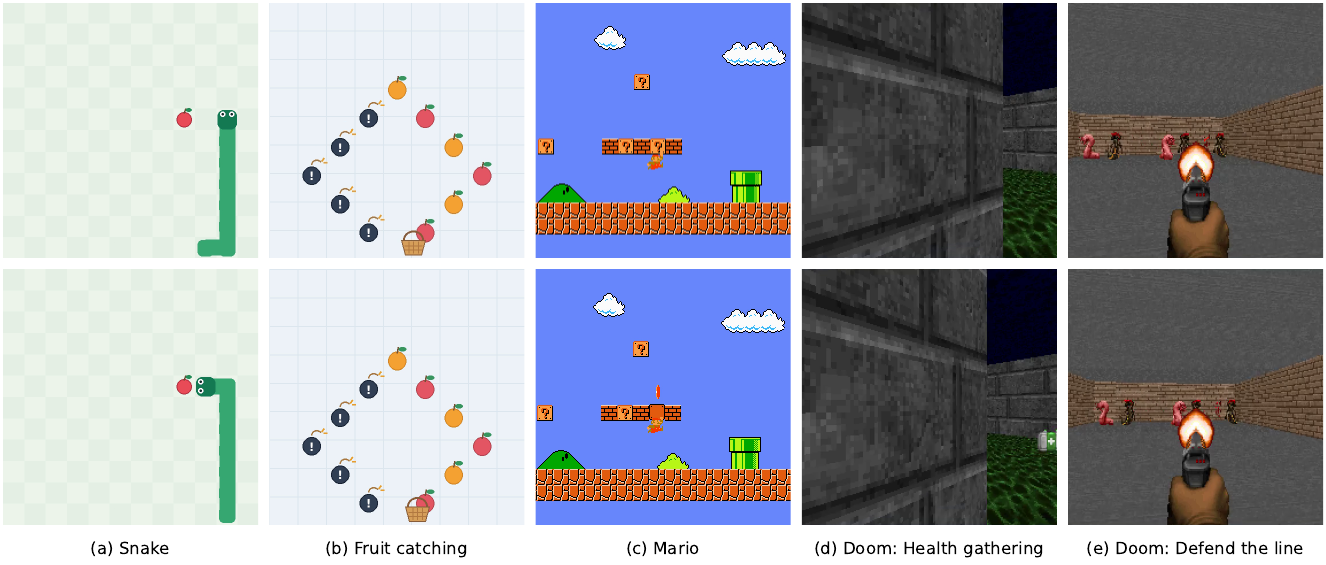}
\caption{Qualitative examples of visual game control with Intern-Decision. From left to right: navigating toward food while avoiding the snake body; catching fruit while avoiding bombs; moving and jumping through a Mario level; searching for health packs; and firing at approaching enemies. Each column shows two consecutive video frames, with the earlier frame on top and the later frame below.}
\label{fig:visual-game-control}
\end{figure}

In the Doom health-gathering example, the agent changes its view after contacting a wall and subsequently searches for new targets. This is an observed behavior in a trajectory; it is not evidence of a learned exploration objective or a general exploration algorithm. Episode counts, seeds, returns, action frequency, observation history, and comparison policies are required for quantitative conclusions about game competence. The short text-request timing in Table~\ref{tab:latency} does not establish visual-control frame rate.

\subsection{Visual selection and mouse actions}
CAPTCHA-style visual tasks require the integration of visual perception, instruction understanding, and interactive decision making. Intern-Decision directly observes interface screenshots, identifies targets described in natural language, and completes tasks through discrete mouse movements and clicks. The model must interpret object attributes and their relationships, translate visual judgments into executable actions, and use the updated screenshot after each action to guide the next decision, forming a closed loop between perception and execution.

\begin{figure}[H]
    \centering
    \includegraphics[width=0.94\textwidth]{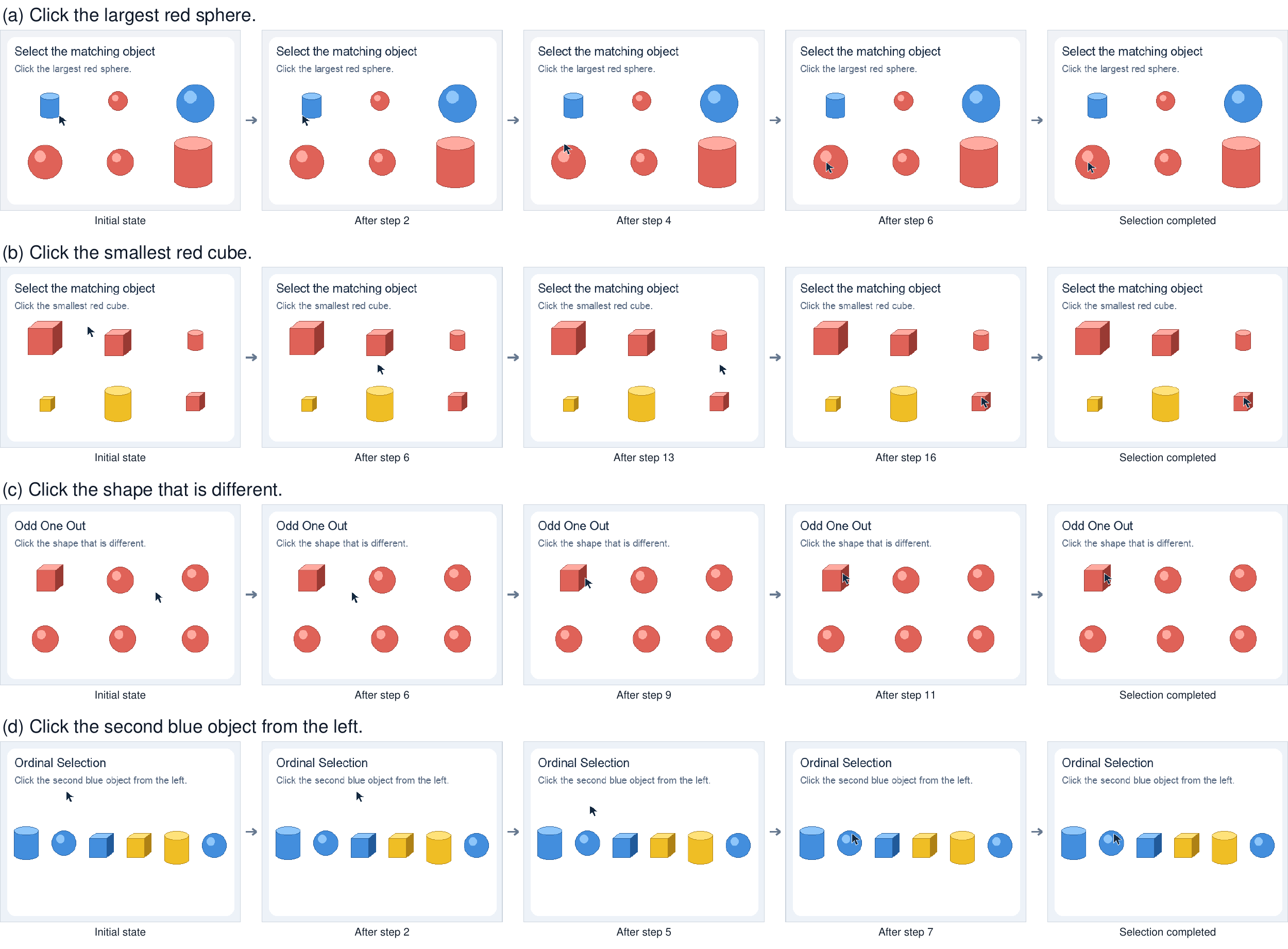}
    \caption{Visual selection trajectories rendered as static video strips. Each row contains five chronologically ordered frames: (a) selecting the largest red sphere; (b) selecting the smallest red cube; (c) identifying the odd shape; and (d) selecting the second blue object from the left. Arrows indicate temporal progression, with intermediate steps omitted. Panels preserve the recorded object layouts and cursor positions; interface text is translated into English for presentation, and video presentation overlays are omitted.}
    \label{fig:visual-selection}
\end{figure}

Four demonstrations record successful trajectories in which Intern-Decision autonomously completes the tasks; Figure~\ref{fig:visual-selection} presents selected keyframes in chronological order. Guided by the task instruction and the current screenshot, the model progressively moves the pointer, adjusts its position, and clicks the target without human takeover or correction. These examples illustrate the integration of multimodal understanding with sequential action selection and provide a concrete experimental setting for studying visual interaction and task execution in graphical interfaces.

\FloatBarrier
\subsection{Hierarchical browser assistance}
\label{sec:browser-use}
We study browser use as hierarchical, closed-loop decision making rather than end-to-end free-form generation. Given a natural-language goal, a larger planner maintains cross-page context and proposes a stage-level subgoal. At each iteration, the current page evidence and a finite candidate set (e.g., click, scroll, return, or stop) are presented to Intern-Decision; the selected action is executed, and the page is observed again. The user task contains neither element identifiers nor a scripted click sequence, so local actions are chosen from the live interface.

\begin{figure}[H]
    \centering
    \includegraphics[width=\linewidth]{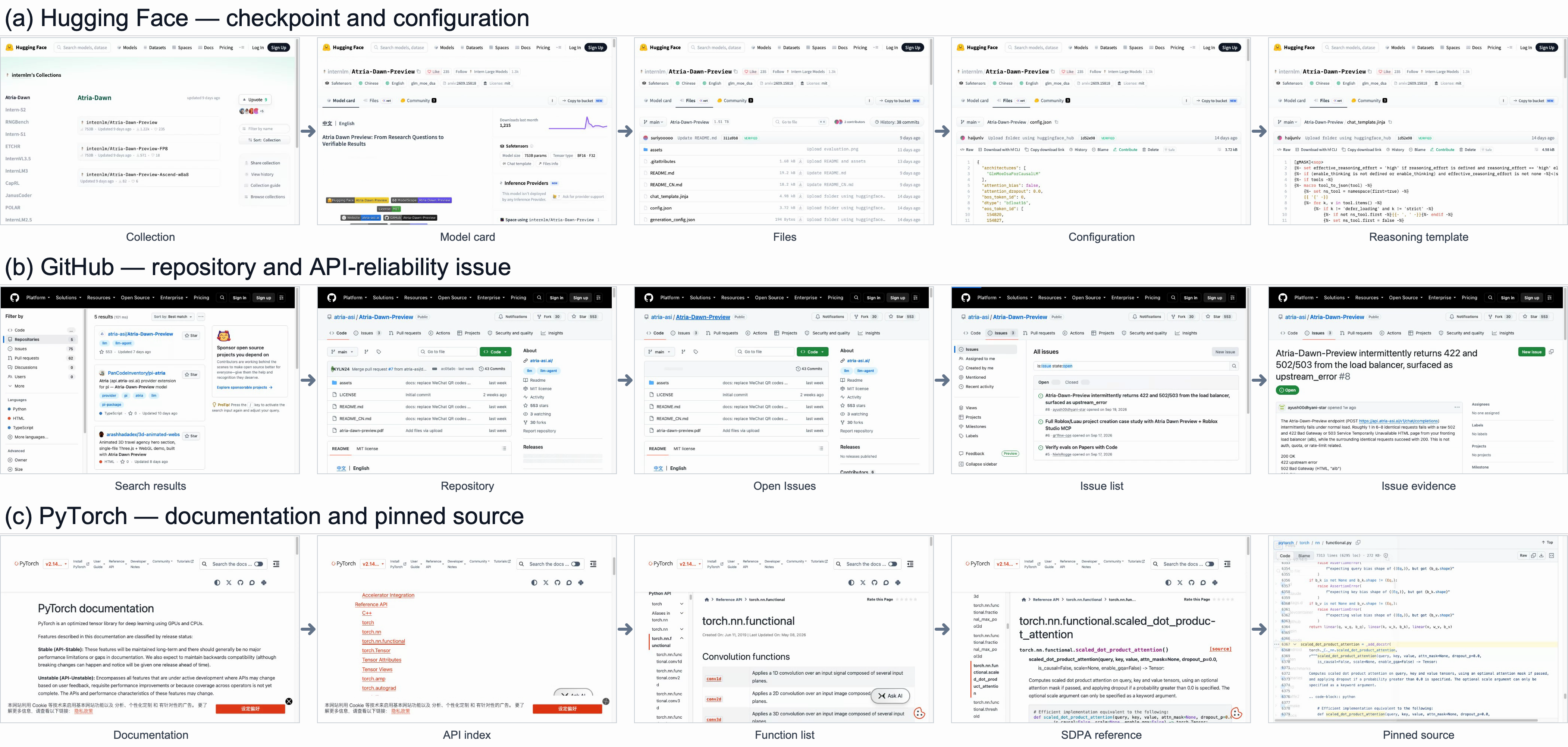}
    \caption{Browser-use trajectories rendered as static video strips. Each row contains five chronologically ordered frames: (a) Hugging Face checkpoint and configuration verification; (b) GitHub repository and API-reliability investigation; and (c) PyTorch documentation and pinned-source inspection. Arrows indicate temporal progression. All panels show the browser interface itself without the presentation overlays used in the demo videos.}
    \label{fig:browser-use}
\end{figure}

Figure~\ref{fig:browser-use} shows three selected trajectories on public websites. The Hugging Face task identifies the standard Atria-Dawn-Preview checkpoint and verifies both its configuration and default-reasoning template. The GitHub task starts from repository search, enters the official project, and locates an API-reliability issue reporting intermittent HTTP 422 and 502/503 responses. The PyTorch task verifies the evaluation-time dropout warning for \texttt{scaled\_dot\_product\_attention} and follows the documented API to its pinned v2.14 implementation. The three traces reach the requested evidence in 6, 6, and 4 local decisions, respectively. Median Intern-Decision inference latency is 101.6, 109.8, and 95.3 ms (server-reported inference time, excluding planning, network, and browser execution).

These examples are qualitative system demonstrations rather than a browser benchmark. Dynamic pages can occasionally produce delayed or visually unchanged outcomes; the controller therefore re-observes the interface after every action and uses explicit completion checks to avoid unnecessary continuation. This division of labor lets the larger model handle goal decomposition and cross-page reasoning while Intern-Decision handles repeated, structured action selection.
\FloatBarrier

\section{Conclusion}
We introduce Intern-Decision, which makes finite multimodal choices through a single causal forward pass, retaining the pretrained backbone's native language-model head and exposing candidate probabilities through a typed interface. The 4B model achieves the highest seven-suite average in the presented comparison, with a 44.16\,ms mean for a fixed short request on an RTX 4090. Separate calibration and handoff studies show how probabilistic outputs can support uncertainty analysis and selective use of a larger model. We believe Intern-Decision can serve as a practical component of larger agent systems, but its generalization to arbitrarily large schemas and complex environments remains an open problem.

\section*{Acknowledgments}
This work is supported by Shanghai Artificial Intelligence Laboratory. Besides, we would like to appreciate Houquan Zhou for the valuable comments on calibration evaluation set construction.

\FloatBarrier
\bibliographystyle{plainnat}
{\small\raggedright\setlength{\bibsep}{2pt}\bibliography{refs}}
\clearpage
\appendix
\section{Release and Reproduction Details}
\label{app:reproduction}
The public implementation documentation inspected for this draft is fixed to Git revision \href{https://github.com/InternLM/Intern-Decision/tree/2f815802058b2464144218859ea9221c1bc0d2a8}{\texttt{2f815802058b}}. It identifies ordinary next-token alignment for the masked skeleton; historical same-position masked experiments are separate revisions. The published accuracy and pilot tables use XTuner, while the reported local latency uses native Hugging Face inference. A checkpoint, tokenizer, processor, calibration artifact, and backend configuration together define an evaluated system.

The released evaluation procedure consists of four stages. First, verify the bundled test hashes and counts. Second, bind the correct preset temperature to the exact checkpoint through metadata and weight-shard checks. Third, run uncalibrated inference and replay the fixed temperature on the saved probabilities, retaining complete raw outputs and verifying unchanged labels. Fourth, score the pilot separately using its exact references and require 96/96 valid outputs. External-baseline adapters and historical responses are not included, so reproducing Intern-Decision rows does not automatically reproduce every baseline row.

\begin{table}[htbp]
\centering
\caption{Full-precision released temperature presets. Each value belongs to its checkpoint and the reported XTuner execution setting; it is not a freely transferable model-size constant.}
\begin{tabular}{lr}
\toprule \textbf{Checkpoint family} & \textbf{Temperature} \\ \midrule
Intern-Decision-0.8B & 2.747760550702957 \\
Intern-Decision-2B & 2.100509348277736 \\
Intern-Decision-4B & 1.9924182353655278 \\
\bottomrule
\end{tabular}
\end{table}

\subsection*{Recovered Training Settings}
\label{app:4b-recipe}
Frozen configurations, input audits, actual training logs, and tokenizer caches identify the following common settings for the released 0.8B, 2B, and 4B \texttt{hf-156} checkpoints. Their input files have the same SHA256, \texttt{8d3a687d7f84}\ldots; the objective revision is \texttt{masked-next-token-v4}. Published weight-shard hashes bind each model to its corresponding completed run. The smaller models come from the September 26 size-scaling runs; the 4B model is their completed September 25 reference.

\begin{table}[htbp]
\centering
\small
\caption{Recovered settings shared by the released 0.8B, 2B, and 4B runs. Global batch size counts packed sequences rather than raw examples.}
\begin{tabularx}{\linewidth}{lL}
\toprule \textbf{Setting} & \textbf{Recorded value} \\ \midrule
Initialization & Matching Qwen3.5-0.8B, Qwen3.5-2B, or Qwen3.5-4B backbone \\
Epochs / optimizer updates & 2 / 156 \\
Peak / minimum learning rate & $2\times10^{-6}$ / $2\times10^{-7}$ \\
Schedule / warm-up ratio & Cosine / 0.03 \\
Global batch / sequence length & 16 packed sequences / 8,192 tokens \\
Precision / GPU count / seed & BF16 / 4 / 42 \\
Maximum gradient norm & 1.0 \\
Frozen modules & Vision tower and projector \\
Input audit & 17,880 rows; 23,500 fields; 5,500 image rows; no truncation \\
\bottomrule
\end{tabularx}
\end{table}

\section{Training Data Examples}
\label{app:training-data-examples}

\subsection*{Composition Overview}
\label{app:training-data-composition}

Table~\ref{tab:training-data-components} summarizes the frozen Intern-Decision training mixture. Counts refer to training samples; multi-question records can contribute more than one decision. The total contains 17,880 samples, 23,500 decisions, and 5,500 image-bearing samples.

\begin{table}[htbp]
\centering
\small
\renewcommand{\arraystretch}{1.15}
\caption{Frozen training mixture shared by the released Intern-Decision models.}
\label{tab:training-data-components}
\begin{tabular}{@{}lrr@{}}
\toprule
\textbf{Component} & \textbf{Samples} & \textbf{Decisions} \\
\midrule
Synthetic decision text & 6,711 & 6,711 \\
Public task distributions & 5,669 & 10,789 \\
Synthetic visual and interactive data & 5,500 & 6,000 \\
\midrule
\textbf{Total} & \textbf{17,880} & \textbf{23,500} \\
\bottomrule
\end{tabular}
\end{table}

The synthetic text data were generated by the teacher model \texttt{deepseek-v4.1-flash}. General contains the 4,066 records from general-category synthesis. Hard combines the 1,762 hard-batch and 883 patch-batch records, giving 2,645 records from difficult-category synthesis. These reporting categories follow the synthesis purpose rather than each record's difficulty annotation.

Public task distributions cover AG News, ToolACE, Typed Decision, WildGuard, WildJailbreak, and Ruozhiba. The group name describes task sources, not uniform label provenance: Typed Decision references were generated by the project, and Ruozhiba labels are marked as LLM-assisted. The shared input and all three tokenizer caches contain the 469 Ruozhiba records. Synthetic visual sources contain rendered observations and action labels from nine game environments (500 records each), VizDoom (500 records), and shapes/captcha-style tasks (500 records).

The nine games are 2048, Car Racing, Flappy Bird, Maze, Frozen Lake, Sokoban, Snake, Catch, and Tetris. VizDoom and these game environments supply synthetic training data; they are not evaluation suites in the quantitative comparison. The demonstrations in Section~\ref{sec:browser-use} and the preceding application sections illustrate behavior rather than held-out scores for these environments. Internal partitions used during synthetic-data preparation are distinct from the test sets used to report benchmark results.

\subsection*{Text Decision: Probability Judgment}

\textbf{State.} A bin holds 40 identical tokens; exactly 3 of them are gold. Five tokens are drawn at random, without replacement.\\[3pt]
\textbf{Question.} Using exact combinatorial reasoning, answer whether the probability of drawing at least one gold token exceeds 0.45. A ``yes'' requires the probability to be strictly greater than 0.45; any equality or lower value means no.\\[3pt]
\textbf{Options.} \texttt{no}: the probability is not greater than 0.45; \texttt{yes}: the probability is strictly greater than 0.45.\\[3pt]
\textbf{Target.} \texttt{no}.

This example is from the probability portion of decision data. It requires exact computation rather than matching surface keywords.

\subsection*{Text Decision: Tool Routing}

\textbf{State.} The user asks for information about a farm with ID 4532. The candidate tools are \texttt{getFarm} (``Retrieve farm information'') and \texttt{addCropToFarm} (``Add a crop to a farm'').\\[3pt]
\textbf{Question.} Ascertain the catalog function that retrieves farm information for a given farm ID.\\[3pt]
\textbf{Target.} \texttt{getFarm}.

This example comes from the public task distribution portion and tests whether the model can select the correct tool from a small candidate catalog.

\Needspace{17\baselineskip}
\subsection*{Visual Decision: Snake}
\noindent\begin{minipage}[t]{0.30\linewidth}
\vspace{0pt}
\includegraphics[width=\linewidth]{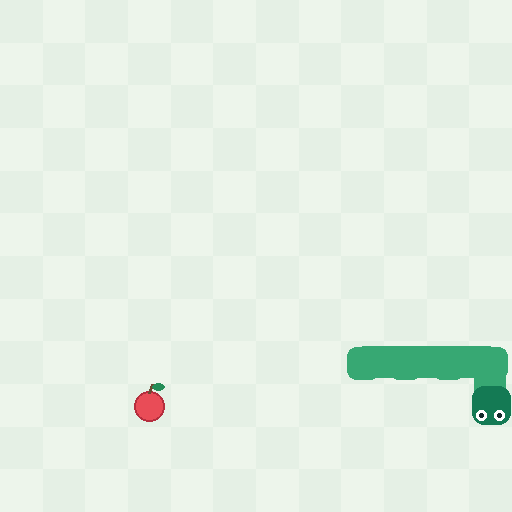}
\captionsetup{font=small,hypcap=false}
\captionof{figure}{A Snake training frame.}
\label{fig:appendix-snake}
\end{minipage}\hfill
\begin{minipage}[t]{0.66\linewidth}
\vspace{0pt}
\textbf{Question.} Choose the next move to guide the snake toward the red apple without hitting a wall or its own body. The end with eyes is the head. The snake grows when it eats and cannot reverse direction.

\textbf{Options.} \texttt{down}, \texttt{left}, \texttt{right}, \texttt{up}.

\textbf{Target.} \texttt{left}.

This example is from the nine game tasks. The model maps a visual game state to a discrete movement decision.
\end{minipage}
\par\medskip

\Needspace{17\baselineskip}
\subsection*{Visual Decision: VizDoom (Two Questions per Frame)}
\noindent\begin{minipage}[t]{0.30\linewidth}
\vspace{0pt}
\includegraphics[width=\linewidth]{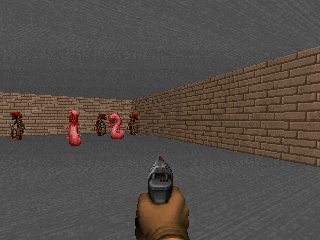}
\captionsetup{font=small,hypcap=false}
\captionof{figure}{Paired turn and attack decisions over a VizDoom frame.}
\label{fig:appendix-vizdoom}
\end{minipage}\hfill
\begin{minipage}[t]{0.66\linewidth}
\vspace{0pt}
\textbf{Question 1 (turn).} Aim toward the largest clearly visible enemy. Choose left or right according to its horizontal position. Stay when it is centered. If no target is visible, search by turning right. \textbf{Target:} \texttt{left}.

\textbf{Question 2 (attack).} Fire when the horizontal screen center lies inside the largest visible enemy's silhouette span. Otherwise aim first. \textbf{Target:} \texttt{no}.

VizDoom records contain two decisions per image, illustrating multi-field supervision over a shared visual state.
\end{minipage}
\par\medskip

\Needspace{17\baselineskip}
\subsection*{Visual Decision: Captcha-Style Shapes}
\noindent\begin{minipage}[t]{0.30\linewidth}
\vspace{0pt}
\includegraphics[width=\linewidth]{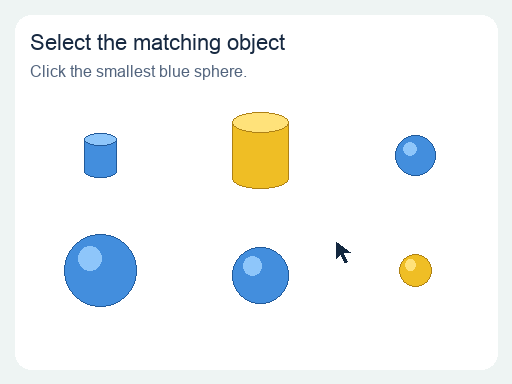}
\captionsetup{font=small,hypcap=false}
\captionof{figure}{A shapes-based training frame.}
\label{fig:appendix-shapes}
\end{minipage}\hfill
\begin{minipage}[t]{0.66\linewidth}
\vspace{0pt}
\textbf{Question.} Based on the task and the current screenshot, choose the next mouse action.

\textbf{Options.} Move the pointer by 10 or 30 pixels in one of four directions, or click at the current pointer tip.

\textbf{Target.} \texttt{RIGHT\_30}.

This example comes from the frozen shapes/captcha-style subset and requires target localization and discrete pointer-action selection.
\end{minipage}
\par

\clearpage
\section{Known-Distribution Test Construction and Cases}
\label{app:calibration-cases}

\subsection*{Mechanism coverage and reference construction}
The pilot consists of six categories with four families each. Every family has two settings and every setting has two option-order variants. The resulting 96 records cover the following mechanisms:

\begin{enumerate}
\item \textbf{Direct randomness and support:} uniform die faces with impossible options; weighted candy inventories; deterministic single-color pens; rare winning lottery tickets.
\item \textbf{Composed events and mixtures:} sums of two dice; numbers of independent hits; independent backup generators; delivery outcomes under an unobserved route mixture.
\item \textbf{History, conditioning, and hidden state:} independent coin flips after a streak; socks after known removals; die outcomes given a restricted sensor observation; a two-state Markov process with noisy sensor readings.
\item \textbf{Daily evidence and observation bias:} defect screening with a stated base rate; conditionally independent repeated tests; correlated copies of one test; waiting times observed at a uniformly sampled time.
\item \textbf{Selective disclosure and puzzles:} informed-host Monty Hall; uninformed-host Monty Hall conditioned on a safe reveal; prize-dependent switch offers; two-child problems with different child-selection protocols.
\item \textbf{Sequential and combinatorial processes:} distinct coupons collected after finite draws; birthday collisions on a toy calendar; gambler's ruin; first-reward times with a guaranteed stopping limit.
\end{enumerate}

The everyday descriptions define synthetic mechanisms with explicit parameters, not empirical frequencies inferred from real-world populations. Their oracles use exact rational arithmetic: finite counting, Bayes' rule and total probability, hidden-state forward updates, coupon-count dynamic programming, birthday no-collision products, gambler's-ruin formulas, and truncated geometric probabilities. An input/reference pair carries a stable ID and input hash. Validation verifies the option keys and probability support, exact normalization, and unchanged references when the option insertion order is reversed. Reference metadata records the family, setting, exact fractions, and derivation separately from the model-visible input.

\subsection*{Support and conditioning: two further actual cases}
\textbf{Fair die with impossible options.} The state specifies a fair six-faced die labeled 1 through 6, rolled once without landing on an edge or leaving the table. The question is ``What number will the die show?'' The ordered criteria contain labels \texttt{1}, \ldots, \texttt{8}. The reference map is
\[
q=\{1\!:\!1/6,\;2\!:\!1/6,\;3\!:\!1/6,\;4\!:\!1/6,
\;5\!:\!1/6,\;6\!:\!1/6,\;7\!:\!0,\;8\!:\!0\}.
\]
This tests equal mass across possible outcomes and zero mass on impossible ones. Returning one face with probability one fails the distribution test. Even a perfect prediction has expected Brier $5/6$, owing to outcome randomness, but zero excess expected Brier.

\textbf{Sampling after removal.} The state says a drawer initially contained six red and four blue socks; exactly three red socks were removed without replacement, with no other changes. One remaining sock is drawn uniformly. The question is ``What color is the selected sock?'' The reversed-order record lists \texttt{blue} then \texttt{red}; its exact reference is \texttt{blue}: $4/7$, \texttt{red}: $3/7$. The model must update the inventory before predicting the draw. Reversing criteria changes neither event probabilities nor the labels attached to them. Both cases use the same common random-experiment instruction shown in the Monty Hall input below.

\clearpage
\subsection*{Monty Hall: prize location as the outcome}

Here is an example of the three-door problem and models' predictions.

\begin{lstlisting}[frame=single,rulecolor=\color{decisionblue}]
{
  "state": "A game has 3 doors and one prize, placed uniformly. You select door 1 before receiving any information. A host who knows the prize location always opens exactly 1 unchosen doors that contain no prize, then always offers a switch to the only other unopened door. If more than one legal reveal set exists, the host chooses uniformly among them. You observe this protocol being completed, but are not told which door numbers were opened.",
  "questions": {
    "decision": {
      "type": "choice",
      "instructions": "Where is the prize relative to the final two unopened doors? Predict the outcome of the specified random experiment using only its stated rules. The listed outcomes are mutually exclusive and exhaustive. Treat every stated probability, independence assumption, and observation protocol as exact.",
      "criteria": {
        "initial_door": "initial_door",
        "other_unopened_door": "other_unopened_door"
      }
    }
  }
}
\end{lstlisting}

\par\smallskip
\noindent\makebox[\linewidth]{\includegraphics[width=0.95\linewidth]{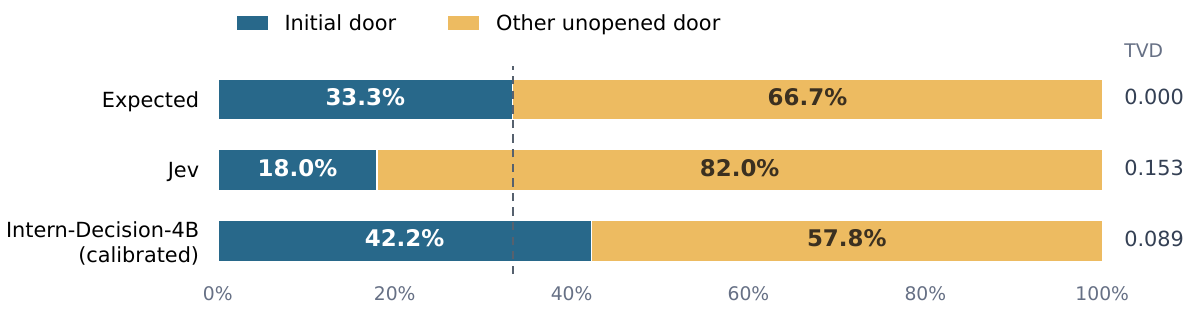}}
\par\smallskip

\end{document}